# Propagation of Delays in the National Airspace System

**Kathryn B. Laskey, Ning Xu, Chun-Hung Chen**
Center for Air Transportation Research and SEOR Department, George Mason University
4400 University Drive
Fairfax, VA 22030-4400
[klaskey, nxu, cchen9]@gmu.edu

## Abstract

The National Airspace System (NAS) is a large and complex system with thousands of interrelated components: administration, control centers, airports, airlines, aircraft, passengers, etc. The complexity of the NAS creates many difficulties in management and control. One of the most pressing problems is flight delay. Delay creates high cost to airlines, complaints from passengers, and difficulties for airport operations. As demand on the system increases, the delay problem becomes more and more prominent. For this reason, it is essential for the Federal Aviation Administration to understand the causes of delay and to find ways to reduce delay. Major contributing factors to delay are congestion at the origin airport, weather, increasing demand, and air traffic management (ATM) decisions such as the Ground Delay Programs (GDP). Delay is an inherently stochastic phenomenon. Even if all known causal factors could be accounted for, macro-level national airspace system (NAS) delays could not be predicted with certainty from micro-level aircraft information. This paper presents a stochastic model that uses Bayesian Networks (BNs) to model the relationships among different components of aircraft delay and the causal factors that affect delays. A case study on delays of departure flights from Chicago O'Hare international airport (ORD) to Hartsfield-Jackson Atlanta International Airport (ATL) reveals how local and system level environmental and human-caused factors combine to affect components of delay, and how these components contribute to the final arrival delay at the destination airport.

## 1 INTRODUCTION

A great deal of research attention has been devoted to the study of flight delay. Traditional linear or nonlinear regression methods have been applied to explain the influence of causal factors on delays. Micro- and macro-level simulation tools have been applied to simulate delays at different levels of detail. Over the years, research methods have shifted from independently investigating particular components of delay, to simultaneously examining multiple components of delay within a single analysis. Examining different components of delay together is important because the components interact in complex ways under the effects of airport conditions, weather conditions, and system effects from NAS. However, "our ability to predict delays because of weather has not improved." And "system predictability in convective weather remains an unresolved puzzle." [1]

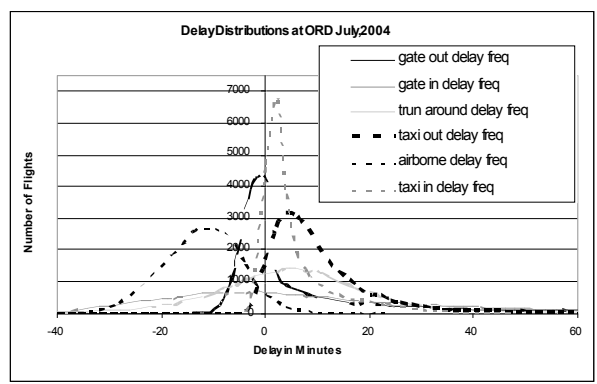

Figure 1: Marginal Distribution of Delay Components for Flights Departing from ORD in July, 2004

Figure 1 shows the marginal distributions of different components of delay in July of 2004. Gate out delay, turn around delay and taxi out delay refer to delays occurring on the ground at ORD. Gate in delay and taxi in delay refer to delays at the destination airports in the NAS when the origin airport is ORD. Airborne Delay refers to the delay in the air between ORD and the destination airport. The curves in this figure are based on 26372 records from the ASPM database for flights leaving ORD in July of 2004. The probability density function was estimated by discretizing the continuous data into 2-minute bins.

Examining just the marginal distributions does not reveal the effects of weather or airport conditions (e.g., congestion), nor does it reveal the relationships of the components in the figure to each other. The Bayesian network model presented in this paper goes beyond the marginal distribution, providing a methodology for quantitatively analyzing the major causal factors affecting each delay component and the relationships among the delay components. The Bayesian network model not only provides predictions of future delays that incorporate the interrelationships among causal factors, but also provides a means of assessing the effects of causal factors and inferring the factors that contributed most to the final arrival delay.

We choose ORD as the departing airport because about 70% of departures from ORD are connecting flights. Both ORD and ATL are listed among the airports with the most serious delay problems. Since delays in different flight phases tend to have different causes, one BN model segment was developed for each phase of flights from ORD to ATL. In each phase, candidate causal factors were identified by examining the literature, and final choices were made through regression analysis. The resulting regression model was input into the BN model segment as a prior distribution for its respective delay node. Next, these segments were linked together into a complete BN model. Finally, historical data was used to estimate a posterior distribution for each node in the BN model.

The remainder of the paper is organized as follows. In Section 2, we describe the data used to build our models. Section 3 presents the methodology to develop BN model segments and describes the complete BN model constructed from all segments. In Section 4, we evaluate our approach and analyze the predictive performance of the model. Section 5 summarizes our results. Section 6 presents conclusions and suggests directions for future research.

## 2 DATA ON FLIGHT DELAY

The data used in this research comes from the Aviation System Performance Metrics (ASPM) and National Convective Weather Detection (NCWD) databases. In the ASPM database, early departures and arrivals were assigned zero delay. In order to incorporate more detailed information about each component of delay, we computed negative delay values for flights that arrived earlier than scheduled. A new database was constructed that combines data from the two sources and includes the computed negative delay variables. Records in the constructed database are indexed by aircraft tail number, and contain information for each variable in our model.

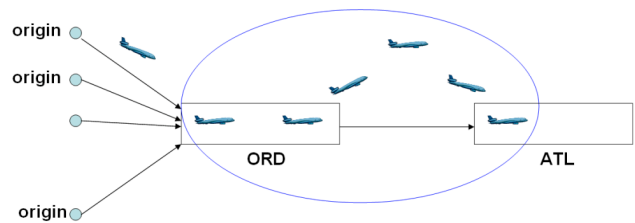

Figure 2: Scope of this Paper.

The delay variables used in this paper are defined as follows:

GateInDelay(ORD, origin): Gate in delay is the difference between the actual gate in time at ORD and the scheduled gate in time from the origin airport.

TurnAroundDelay(ORD): Turn around delay is the difference between the scheduled turn around time at ORD and the actual turn around time.

GateOutDelay(ORD,destination): Gate out delay is the difference between the actual gate out time at ORD and the scheduled gate out time to a destination airport.

TaxiOutDelay(ORD): Taxi out delay is the difference between the actual taxi out time at ORD and the unimpeded taxi out time.

AirborneDelay(ORD,destination): Airborne Delay is the difference between the actual airborne time from ORD to a destination airport and the predicted airborne time in the flight plan. For positive delays, this is the airborne delay defined in ASPM; for negative delays, the value in ASPM is zero.

TaxiInDelay(destination): Taxi in delay is the difference between the actual taxi in time at the destination airport and the unimpeded taxi in time.

GateInDelay(destination,ORD): Gate in delay is the difference between the actual gate in time of flights from ORD at a destination airport and the scheduled gate in time.

Throughput: Throughput is measured by counting departures or arrivals in 30-minute windows or in 15-minute windows for different flight phases.

To incorporate factors from air traffic management, we derived some variables from the ASPM database to represent effects of the ground delay program (GDP). When GDP is in effect, the ASPM field EDCTOFFSEC (EDCT wheels off second) represents the predicted earliest time for the aircraft to be released for takeoff. The EDCTOFFSEC assignment is a way for air traffic controllers to keep a given flight from arriving at the destination airport at the scheduled time. When GDP is not in effect, the value of EDCTOFFSEC is -1. Three new variables are derived for the impact of the ground delay program in our model. They are GDP, GDPtime and GDPgate. GDP is a Boolean variable set to false if EDCTOFFSEC=-1 and true otherwise. The variable GDPtime is defined as the difference between the flight's actual pushback time (ACTOUTSEC in seconds) and the ETMS planned pushback time (NOMTO in minutes) assuming there is no taxi out delay. When GDP is false, it has value zero; otherwise its value is

$$GDPtime=[EDCTOFFSEC-(ACTOUTSEC+NOMTO*60)]/60$$

If GDPtime is greater than zero, that means the aircraft should have taken off earlier than the ETMS assigned takeoff time if it did not have to wait in a departure queue. This further implies that GDP causes its taxi out delay. If GDPtime is less than zero, then GDP did not affect the current flight's takeoff process, although it might have affected the pushback process if the aircraft was held at the gate because of GDP.

## 3 METHODOLOGY

From departure to arrival, an aircraft pushes back from the gate, taxis out to the runway, takes off, passes though many en route sectors in the air, lands, and finally taxis to the gate. At the gate, the aircraft waits for turn-around, after which it continues on to the next leg. Previous research has identified that each phase of flight has different sensitivities to weather [2]. Many articles have analyzed the effects of demand and capacity together with weather for a given phase of flight delay [5,6,7,8]. In this paper, we decompose the final arrival delay into different parts according to the phase of flight and develop BN model segments for the delay in each phase.

These BN segments are then linked together through the common variables in each segment to construct a BN model for all phases from the time a flight turns around at ORD until it arrives at the gate at ATL. Data from July to September 22nd in 2004 were used to develop the BN model and estimate its parameters. Data from the last week of September was withheld to test the model's prediction accuracy. These two sets of data are called the training and test samples, respectively. There were 2019 cases in the training sample and 146 cases in the test sample. This is a small sample for estimating a model of this complexity.

A regression model was constructed and evaluated for each phase of delay. The dependent variable was the delay at the given phase. The potential independent variables were delays from previous phases and other explanatory variables identified from the literature. The model construction process proceeded as follows. For each phase of delay, the following steps were performed:

1. Distinguish the most important explanatory factors for this phase using piece-wise regression analysis and cross validation on the training sample.

2. Create a node in the BN to represent the delay phase.

3. Set the factors selected from step 1. as the parent nodes of the given delay node in the Bayesian network.

4. Estimate initial local distributions for the given node by discretizing the regression model. That is, the child node is modeled as a normal distribution with mean equal to the regression mean and standard deviation equal to the regression standard deviation. Most delay variables were discretized in 15-minute intervals, but some were discretized more finely to improve accuracy.

5. Use Dirichlet-multinomial learning from the training data to update the distributions of all nodes in the Bayesian network. We found this step to be necessary because the regression model alone was not adequate to capture the complex relationships between nodes and their parents. We gave a relative weight of 30:1 on observed cases to the regression prior (see discussion in Section 4 below).

6. Evaluate the model by comparing the model predictions with observations on a holdout sample.

After distributions were constructed for each phase of delay, the BN model segments were combined into the Bayesian network shown in Figure 3. The dark boxes in Figure 3 represent factors that affect delay in the given phase. Each of these corresponds to a set of nodes in the final Bayesian network. We used Belief Network Power Constructor to identify arcs between the causal factors. These arcs are not shown in the figure. The node, GateInDelay(ORD,origin,tail#), has a different color from the other delay variables because it is treated as an input and its causal factors are not being modeled here (although we plan to repeat this modeling process for flights into ORD, at which time this node will be treated as a dependent variable). All delay variables point directly or indirectly to the gate arrival delay at ATL, GateInDelay(ATL,ORD,tail#). That is, the model provides an estimate of the probability distribution for the gate in delay at ATL conditional on the gate in delay at ORD and the causal factors depicted in the dark boxes.

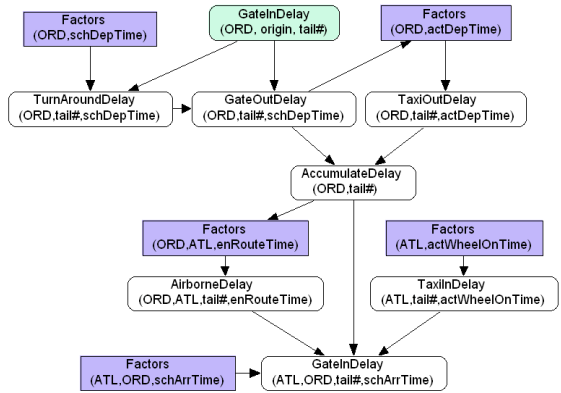

Figure 3: Structure of BN Model for All Phases

Table 1 shows the parents of each random variable in our model. Identifiers for each random variable are included in parentheses. We omit the tail number in each variable to save space because it appears in all variables listed in Table 1. In Table 1, *act* represents actual, *sch* represents scheduled, *Arr* means arrival, and *Dep* means departure.

## 4 MODEL EVALUATION

A common model evaluation metric is mean squared error (MSE) on training and test samples. We built our models using Netica, which has limited support for continuous random variables and does not provide MSE values. We calculated an approximate MSE from the confusion matrix output by Netica as follows.

TABLE 1
PARENT NODES OF DELAY VARIABLES

| Delay Variables | PARENT NODES |
|---|---|
| TurnAround Delay(ORD) | GateInDelay(ORDfromOrigin), GDPgate Airline SchGateOutTime, ArrThroughput(ORD), Weather(ATL,schDepTime, 1hrLater) |
| GateOutDelay(ORD) | TurnAroundDelay(ORD), GateInDelay(ORDfromOrigin) |
| TaxiOutDelay(ORD) | DepQueueSize(ORD), ArrivalThroughput(ORD), ActGateOutTime, RunwayConfiguratoin(ORD), EnRoutestrom(ORD,ATL,4hrLater) |
| DepQueueSize(ORD) | GDPtime, ActDepDemand(ORD), ArrThroughput(ORD), Airline |
| AirborneDelay (ORDtoATL) | PredictedEnRouteTime(ORDtoATL), Weather(ATL,actDepTime,3hrLater), ArrThroughput(ATL), EnRoutestrom(ORDtoATL) |
| PredictedEnRouteTime (ORDtoATL) | Airline, AccumulatedDelay(ORDtoATL), EnRouteThunderstorm(ORDtoATL,1hrLater), Weather(ATL,actDepTime,2hrLater) |
| TaxiInDelay(ATL) | DepQueueSize(ATL), ArrQueueSize(ATL), DepThroughput(ATL), ArrThroughput(ATL) |
| ArrQueueSize(ATL) | DepQueueSize(ATL), DepThroughput(ATL), ArrThroughput(ATL), ArrDemand(ATL), ActWheelOnTime(ATL) |
| GateInDelay (ATLfromORD) | AccumulatedDepDelay(ORDtoATL), AirborneDelay(ORDto,ATL), TaxiInDelay(ATLfromORD), Airline, ScheduledGateInTime(ATLfromORD), DepThroughput(ATL) |

$$MSE = \sum (y_i - \hat{y}_i)^2 / n \quad (1)$$

where:

$y_i$ – the midpoint of resident bin of actual delay;

$\hat{y}_i$ – the midpoint of resident bin of estimated delay with maximum likelihood;

$n$ – the number of cases in the testing sample

Table 2 shows approximate MSE values for each of the random variables in our model. As expected, MSE values are generally about the same or slightly worse on the test sample than on the training sample. Performance was better on the model plus Dirichlet-multinomial learning than on the regression model alone.

TABLE 2
COMPARISON OF MSE ON TRAINING AND TEST SAMPLES

| Delay Variables | Training Sample mean (minute) | Training Sample std | Approx. MSE on Training Sample | Approx. MSE on Test Sample |
|---|---|---|---|---|
| TurnAroundDelay | 3.68 | 30.52 | 267.2 | 372.0 |
| TaxiOutDelay | 12.93 | 17.70 | 59.8 | 77.6 |
| AirborneDelay | 1.40 | 11.20 | 86.9 | 103.9 |
| TaxiInDelay | 2.90 | 4.35 | 39.1 | 37.1 |
| GateInDelay | 13.95 | 39.89 | 170.4 | 195.2 |

We experimented with different values of the weight given to the regression prior distribution versus the cases (this weight is also known as the virtual sample size). As expected, heavier weights on cases produced lower approximate mean squared errors on the training sample. On the test sample, the mean squared error first improved with increasing virtual sample size and then worsened. The value that produced the best mean squared error was different for different phases of delay. We chose a weight of 30:1 on prior versus cases as a compromise point that performed well for all phases. Figure 4 shows the relationship of virtual sample size and scaled mean square error for training and testing samples, where:

$$Scaled(MSE_i) = [MSE_i - \min(MSE)] / [\max(MSE) - \min(MSE)] \quad (2)$$

In this equation, $i$ is the case weight and $MSE_i$ is the approximate MSE of model updated using i:1 weighting.

We also compared our model to Dirichlet-multinomial learning alone. The multinomial-Dirichlet model had dramatically better scores on the training sample and dramatically worse scores on the test sample, a clear indication of overfitting to the training sample.

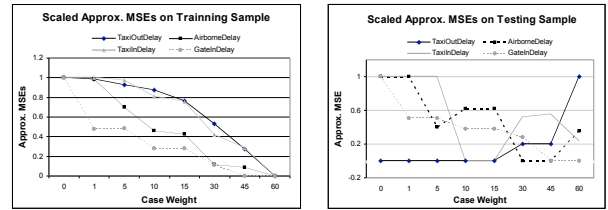

(a) Training sample  (b) Testing sample
Figure 4: Comparison of MSEs for Different Values of Virtual Sample Size

As another model diagnostic, we converted our models into IET's Quiddity*Suite BN tool and computed the squared difference between each node's observed value and its conditional mean given its Markov blanket, for all nodes in the BN and all cases in the test sample. In this evaluation, combining Dirichlet-multinomial and regression out-performed regression alone by a large margin, and equal weighting of prior and observations performed slightly better than the 30:1 weighting.

Finally, we computed log-likelihood values for each of the cases in our holdout sample and compared them with log-likelihood values for a sample randomly generated from our model. If the model were an accurate reflection of the process generating the observations, we would expect these two distributions to be the same. Results are shown in Figure 5 below. It is clear that the two distributions are different. The randomly generated cases have overall lower log-likelihood, and the actual cases have a long lower tail. This result is not surprising because our evaluation uses point estimates of the parameters. A more sophisticated posterior predictive analysis [9] would explicitly consider parameter uncertainty. The distributions of Figure 5 are much more similar than the corresponding distributions from regression only, Dirichlet-multinomial learning only, and also more similar than those from regression plus Dirichlet-multinomial with a 1:1 weight on training sample and prior distribution.

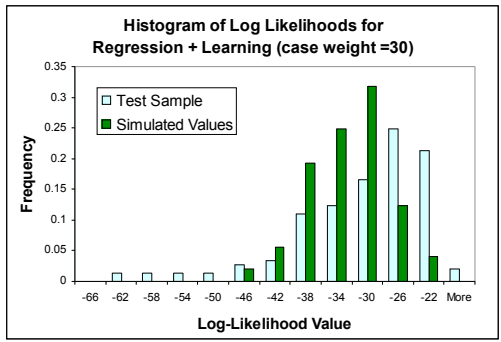

Figure 5: Log-Likelihood Comparison

## 5 ANALYSIS RESULTS

### 5.1 INFERRING CAUSES OF DELAY

A BN model can perform many different kinds of "what-if" queries, in which we set the value of some variables and obtain updated distributions on other variables. In this way, we can investigate how different system and environment variables interact to cause or mitigate delay.

For example, we can specify a value for the gate in delay at ATL, and use the BN to infer the expected value and distribution of other delay variables. This analysis can help to identify the critical phases in the flight itinerary that contribute the most to gate in delay. Figure 6 shows an analysis in which the gate in delay in ATL was varied from 0~15 minutes to more than 120 minutes. The posterior distribution of each of the other delay components given gate in delay at ATL was plotted for the following scenarios:

a) Gate in delay at ATL is between 15 to 30 minutes
b) Gate in delay at ATL is between 60 to 75 minutes
c) Gate in delay at ATL is between 105 to 120 minutes

The plots in Figure 6 show that when arrival delay at ATL is greater than 1 hour, there is a substantial increase

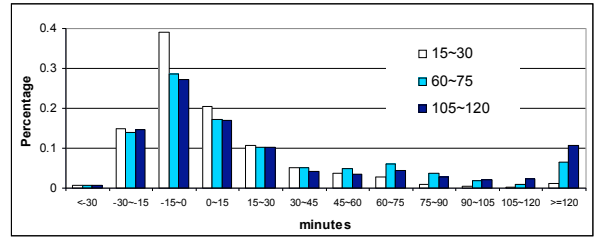
*Posterior distributions of gate in delay at ORD from previous airport*

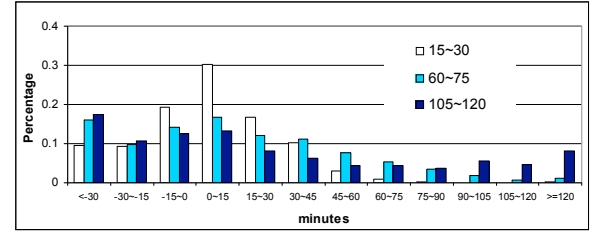
*Posterior distributions of turn around delay at ORD*

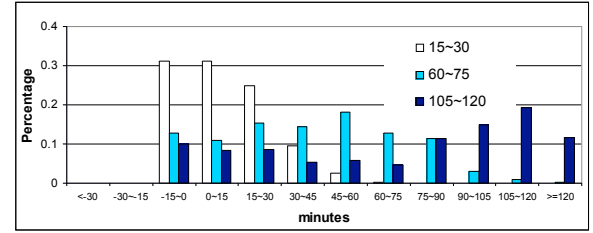
*Posterior distributions of gate out delay at ORD*

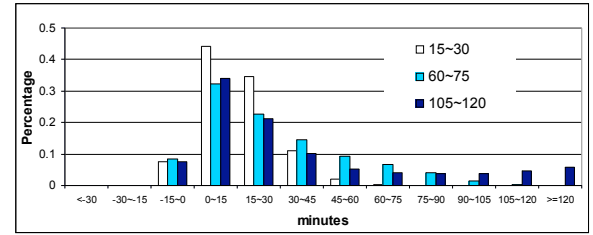
*Posterior distributions of taxi out delay at ORD*

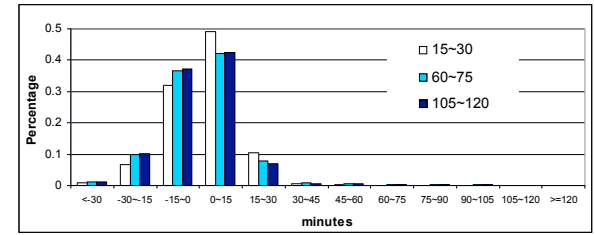
*Posterior distributions of airborne delay from ORD to ATL*

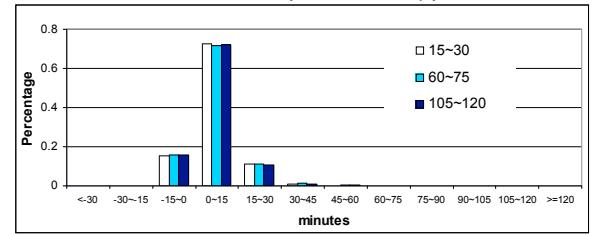
*Posterior distributions of taxi in delay at ATL*

Figure 6: Posterior Distributions of Delay in Each Phase Given 3 Different Values of Gate in Delay at ATL

in the probability that the gate out delay and taxi out delay at ORD are greater than 1 hour. Arrival delays at ATL of longer than 2 hours tend to be associated with long gate out delay at ORD, while the probability increases that turn around delay at ORD is negative. In any scenario, the airborne delay and taxi in delay at ATL do not change much with changes in arrival delay at ATL. Out of 2019 records form ORD to ATL, there are only 45 records having taxi in delay longer than 15 minutes, and among them 4 records having taxi in delay longer than 30 minutes. Since Airborne Delay is computed by the difference between actual en route time and the predicted en route time in the flight plan, the small and constant expected value and associated variance reflect the FAAs' efficiency in en route time prediction. In the next section, we describe the important factors for each phase.

### 5.2 ANALYSIS OF EACH PHASE

*a. Turn around phase*

Figure 7 shows how previous leg gate arrival delay and turn around delay contribute to delay in departing from the gate. Gate departure delays were set at 0~15 minutes (which is counted as on time departure), 45~60 minutes and 75~90 minutes. For on time departing flights, the previous arrival was more likely to be early and turn around delay was more likely to be on time. When gate departure delays were between 45~60 minutes, more than 25% of the flights had gate arrival delay more than 60 minutes and 95% of them made the turn around process faster than the scheduled time. When gate departure delays were between 75~90 minutes, 40% of the flights had arrival delay from the previous leg of more than 1 hour and 60% of the flights decreased the turn around time by more than half an hour.

Figure 7 (b) and (c) show that there is some probability of on time ORD arrival corresponding to a long ORD departure delay. Further analysis shows that the destination airport ATL weather condition, especially low visibility and en route thunderstorm at the scheduled departure time, are major contributing factors to gate departure delay in this circumstance. For instance, when the flights arrived in gate on time, but had more than 1 hour turn around delay, the probability of good weather conditions at ATL dropped from 93% to 81%. When weather was clear at ATL at the flight's scheduled departure time and there was no ground delay program issued, 74% of the flights departed on time and 64% even departed earlier than schedule. When ATL experienced low visibility in the afternoon, only 34% flights from ORD to ATL made on time departure.

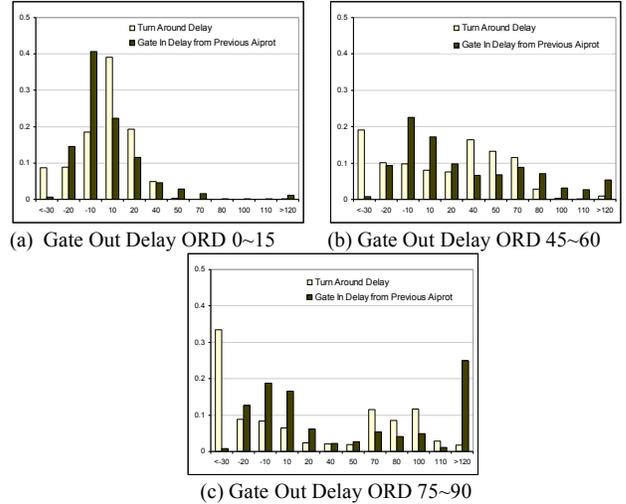

Figure 7: Posterior distributions of gate arrival delay in previous leg and turn around delay given current gate departure delay. (*The x axis is gate departure delay in minutes and y axis is the percentage of two attribute delays.*)

*b. Taxi out phase*

At the taxi out phase, we performed analysis on actual taxi out delay using the factors from existing research [3] via the combination of traditional statistical analysis and our Bayesian Network model. We found airport arrival throughput, departure time, and GDP related variables have considerable influence on taxi out delay in addition to other factors such as airline, departure runway, departure demand [4], and downstream restrictions.

The correlation between taxi out delay and these explanatory factors is 0.91, while the correlation with departure queue size alone is more than 0.8. In our data, the queue size ranged from 0 to more than 140.

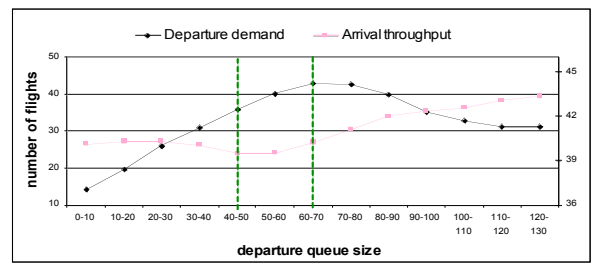

Figure 8: Expected value of posterior distributions of departure demand and arrival throughput given number of queue size.

Figure 8 shows that when the queue size is below 40, departure queue size shows a positive relationship to departure demand and slightly negative relationship to arrival throughput. When the departure queue size grows to between 40 and 70 flights, there is a positive relationship to departure demand and arrival throughput. When there are more than 70 flights in the queue, the queue size has positive correlation to arrival throughput but negative correlation to the departure demand. This phenomenon can be explained by the tradeoff between arrivals and departures of limited airport capacity.

We investigated the cases in which queue size was above 60, which represents 7% of the data in the summer of 2004. As shown in Figure 8, out of 165 records in the database, the percentage of flights having a long GDP holding time increased rapidly when the queue size was above 70 and was close to 100% when the queue size was above 110, a level associated with average taxi out delays of more than 1 hour. This result indicates that long departure queue sizes are highly associated with GDP issuance. The flights not related to the ground delay program might be affected by mechanical problems or other unknown reasons.

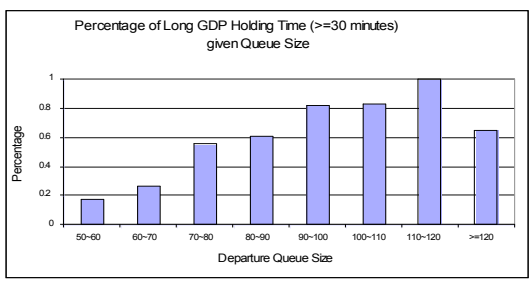

Figure 9: Probability of GDP Holding Time ≥ 30 Minutes For Different Queue Sizes

*c. Airborne phase*

The above analyses demonstrate that airlines make up lost time by speeding up the turn around phase. Similarly, the en route phase also gives airlines some room to make up lost time when they are behind schedule. The average actual en route time compared to the scheduled en route time is -12 minutes, which strongly suggests a tendency of airlines to pad the schedule. In order to investigate schedule padding, we analyzed the difference between the actual en route time and the predicted en route time from the flight plan. We define this difference as Airborne Delay. The average value of this variable is 1 minutes and the variance of this variable is relatively small compared to other delay variables.

The average airborne time between ORD and ATL in summer is 89 minutes. When facing the same amount of accumulated delay, different airlines have different strategies for estimating airborne time. The largest difference in predicted airborne time between major airlines is 10 minutes. Heavy rain and high wind at the destination airport and severe en route thunderstorms can increase the average predicted airborne time of major airlines at ATL to 110 minutes. Under the same situation other airlines (not the major airlines at ATL) did not adjust the prediction airborne time, the probability of having more than 15 minutes airborne delay is 47%.

# 6   CONCLUSIONS

Bayesian networks provide a powerful modeling tool for investigating the causal factors contributing to delay in each flight phase and analyzing the contribution of each phase to the final arrival delay. Our results on a holdout sample demonstrate the prediction ability of the BN model. Backward analysis on attribution of delays shows that departure delays at the busy hub airport ORD are major contributors to the final gate arrival delay at the destination airport. Weather conditions en route and at the destination airport ATL have an effect on delay in all flight phases. Arrival delays longer than 1 hour at ATL are usually associated with severe weather and consequent GDP issuance.

The paper models the distribution of delays between two top delay airports in the summer of 2004. Different conclusions for winter season or different origin-destination pairs would be expected. We plan to continue our research with data from additional airports and time periods.

There are a number of methodological issues we intend to address in future work. Further investigation is needed of the effects of discretization. While discretization introduces error, because Dirichlet-multinomial learning with regression priors performed dramatically better than regression alone, it is not clear that finer discretization would provide an improvement. We plan to investigate more sophisticated non-parametric density estimation methods and inference algorithms tailored to continuous distributions. As additional airports and time periods are added, exact inference will become intractable and we plan to apply approximate inference algorithms. Another promising avenue of research is to develop hierarchical Bayesian models that incorporate a temporal dimension and in which individual airports are modeled as drawn from a population with a common prior distribution. We estimated our models using data from the summer of 2004. Changes in how the airlines operate since that time will affect some parts of our model, while other parts should remain relatively stable. Bayesian hierarchical modeling and the modular construction of our models will facilitate adaptation of the models to current conditions.

Our ultimate objective is to provide a tool that will enable planners to run what-if scenarios to investigate the impact of changes in tactical decisions and policies with respect to the ground delay program and decisions to cancel flights, and to investigate how flight scheduling decisions by individual airlines contribute to the propagation of delay in the system.

## Acknowledgments

This work has been supported in part by NSF under Grant IIS-0325074, by NASA Ames Research Center under Grants NAG-2-1643 and NNA05CV26G, by NASA Langley Research Center and NIA under task order

NNL04AA07T, by FAA under Grant 00-G-016, and by George Mason University Research Foundation. We thank Masami Takikawa for converting our Netica models to Quiddity*Suite and computing model evaluation measures.